\documentclass{article}

     \usepackage[nonatbib,preprint]{neurips_2020}

\usepackage[utf8]{inputenc} 
\usepackage[T1]{fontenc}    
\usepackage{hyperref}       
\usepackage{url}            
\usepackage{booktabs}       
\usepackage{amsfonts}       
\usepackage{nicefrac}       
\usepackage{microtype}      
\usepackage{makecell}
\usepackage{xcolor}
\usepackage{wrapfig}
\usepackage{enumitem}
\usepackage{bm}
\usepackage{csquotes}
\usepackage{lipsum}
\usepackage{tabularx}
\usepackage[ruled,vlined]{algorithm2e}
\usepackage{graphicx}
\usepackage[numbers]{natbib}

\title{Weird AI Yankovic: Generating Parody Lyrics}

\author{%
  Mark O.~Riedl\\
  School of Interactive Computing\\
  Georgia Institute of Technology\\
  \texttt{riedl@gatech.edu} \\
}

\begin{document}

\maketitle

\begin{abstract}
Lyrics parody swaps one set of words that accompany a melody with a new set of words, preserving the number of syllables per line and the rhyme scheme.
Lyrics parody generation is a challenge for controllable text generation.
We show how a specialized sampling procedure, combined with backward text generation with XLNet can produce parody lyrics that reliably meet the syllable and rhyme scheme constraints.
We introduce the Weird AI Yankovic system and provide a case study evaluation.
We conclude with societal implications of neural lyric parody generation.
\end{abstract}

\section{Introduction}

Musical parody is a important part of Western society.
It is used as a form of free speech to advance political debate, give flattery, provide mockery, and to entertain.
In the United States of America parody is perceived as an important part of free speech to the point that it is protected by {\em fair use} laws, meaning that music can be copied if the lyrics are changed in a significant manner.
Musical parody takes advantage of familiar tunes to catch one's attention and bring awareness to the message captured in the new lyrics.
The author of this paper likes to motivate their seven-year old son to get ready to bed by making the announcement to the tune of songs he knows.
Is there really any more important use of speech than motivating one's child to go to bed so one can write a paper for arXiv?

Whereas making a song from scratch requires inventing new melodies and beats, song parody allows a person without any music writing experience to select new words that fit an existing syllable and rhyme scheme.
This is not to say that writing good music parody is easy. 
The artist, ``Weird'' Al Yankovic is famous for publishing well-known parodies of famous musicians such as Michael Jackson, Madonna, and Nirvana. 
Online social media platforms such as Twitter and Facebook have made parody more popular, allowing anyone to share parodies about topical themes, such as the showtune parodies by Randy Rainbow that mock politicians, computer science instructors teaching about machine learning topics,\footnote{\textit{ Overfitting Thriller} by Michael Littman and Charles Isbell \url{https://youtu.be/DQWI1kvmwRg}} or university professors singing about teaching remotely during a pandemic to the tune of {\em I will Survive} by Gloria Gaynor.\footnote{\textit{I Will Survive, Coronavirus version} by  Michael Bruening \url{https://youtu.be/CCe5PaeAeew}}

In this paper, I introduce a system that can generate new parody lyrics for existing songs. 
The system does not generate melodies---that is assumed to already exist because there is a song being parodied---nor does the system sing the lyrics it generates.
The new lyrics are textual and meant to be sung by the user to the original tune. 
Figure~\ref{tab:weird} shows an example of original lyrics side by side with a famous human-written parody, and output from our system.
Our system can also produce a karaoke video so one can sing the new lyrics to original music.
I named the system {\em Weird AI Yankovic} to give homage to the greatest parody musician ever.
Also because if a san-serif font is used, it is hard to tell the difference between \textsf{Al} and \textsf{AI}. 
It is best to read the rest of this paper in the voice of Al Yankovic.

\begin{table}[h!]
    \centering
    \footnotesize
    \begin{tabular}{p{1.6in}|p{1.6in}|p{1.6in}}
{\bf {\em Beat It} by Michael Jackson} & {\bf {\em Eat It} by Weird Al Yankovic} & {\bf System output} \\
\hline
They told him don't you ever come around here &
How come you're always such a fussy young man? &
The best part is that each taco contains a small \\
Don't want to see your face, you better disappear &
Don't want no Captain Crunch, don't want no Raisin Bran &
To medium sized piece of sliced chicken nepal \\
The fire's in their eyes and their words are really clear &
Well, don't you know that other kids are starving in Japan? &
I don't think the food in question lasted awhile \\
So beat it, just beat it &  
So eat it, just eat it & 
I promise, just promise \\
\hline
    \end{tabular}
    \caption{Some examples of lyrics parodies. The original lyrics are on the left. Human-written parody lyrics are given in the center. Output from our system is give on the right.}
    \label{tab:examples}
\end{table}

\section{Parody Lyrics as AI Challenge}

In it's most basic form, lyrics parody swaps one set of words that accompany a melody with a new set of words, preserving the number of syllables per line and the rhyme scheme indicating which lines rhyme with each other.
In doing so, the new lyrics are likely to also fit the melody, which remains unchanged, and will be recognizable to hearers.
The rhyme scheme and number of lyrics per line can be viewed as constraints on a language generation task.
Some examples of parody lyrics, human written and algorithmically generated, are shown in Table~\ref{tab:examples}.

AI lyric parody generation is a form of {\em controllable text generation}. 
Many large-scale transformer-based neural language models, such as GPT-2~\cite{Radford2019LanguageMA}, XLNet~\cite{Yang2019XLNetGA}, T5~\cite{Raffel2019ExploringTL}, or even GPT-3~\cite{Brown2020LanguageMA}, are capable of producing fluent language.
However, neural language models predict a sequence of tokens based on a given sequence of tokens.
That's cool---one can provide a prompt and get a reasonable continuation.
However, the generation is not {\em controllable} because one cannot specify any constraints on what is produced,
such as:
the number of words in a sentence, the number of syllables per sentence, or whether certain words rhyme.
The reason that controllability is challenging is because generative language models do not perform look-ahead.
That is, local word choices do not take into account whether it makes it easier or harder to meet constraints that come in to play later, such as a rhyme.

Our instincts as deep learning researchers would be to train a neural language model from scratch, or fine-tune an existing transformer-based neural language model, to produce a given number of syllables and to end lines with a rhyme. 
That would probably work; I dunno, I didn't try that.
Like, that just sounds hard.
Where do I get the corpus? 
How do I label it? 
Do I have to create a new model for each new song with different pattern of syllables or rhymes?
Can I train a general system and prompt it with the pattern?

An alternative approach is to provide a specialized sampling strategy that is sensitive to the constraints of syllable and rhyme.
The dirty secret of neural language modeling systems is that they can be thought of containing three components:
(1)~an encoder that compresses a prompt into a learned representation,
(2)~a decoder that decompresses a representation and produces a distribution over the vocabulary, and
(3)~a {\em sampler} that selects tokens from the distribution.
The simplest sampling strategy is greedy, taking the logit with the highest value.
Other sampling strategies include top-$k$, nucleus sampling (top-$p$)~\cite{Holtzman2020TheCC}, and beam search.
While I find that top-$k$ or top-$p$ work pretty well for most things I want to do, beam search 
can increase the odds that later constraints are met as the width of the beam is increased, though it can become trapped in local maxima.

In this paper I show how a combination of forward generation with a neural language model, backward generation with a neural language model, and a specialized sampling strategy that is sensitive to syllables can produce parody lyrics that meet a set of given constraints regarding number of syllables per line and rhyme scheme between lines.
This paper presents the engineering considerations because I am not sure there are any scientific contributions.
I just wanted to make something that worked, and, frankly, it worked a lot better than I expected. 


\section{The Weird AI Yankovic System}

This section walks through all the various parts of the system.

{\bf Constraints.}
Music parody starts with understanding the syllable and rhyme constraints from the original music.
Constraints should be easily provided by users.
The user provides a {\em scheme}, a list of line specifications where each line is a list of segments $(s_i, r_i, e_i)$ such that $s_i$ is the number of syllables, $r_i$ is an unique numerical identifier such that each segment with the identifier will rhyme, and $e_i$ is an optional parameter to end the segment with a period.
A line can consist of more than one of these segment tuples because of interior rhymes, used frequently in hip-hop, rap, and {\em Hamilton: the Musical}.
The rhyme identifier can be null, signifying that the line (or segment) does not need to rhyme with anything else.
The rhyme identifier can also be a word or phrase, indicating that the generator must use this exact word or phrase.
A {\em rhyme map} keeps track of words that must be rhymed with for each $r$; the rhyme map can be seeded if the user wants certain lines to rhyme with certain words.

\textbf{Context Prompt.}
The context prompt is another user input that provides a word, phrase, or sentence to cue a particular topic. 
For example: ``my favorite food is tacos''.\footnote{The prompt doesn't always have to be true} 
The prompt is provided as an initial input to see the language model generation for the first line of the lyrics (see below). 
After that, each call to the generative language model is seeded with the prompt plus all subsequent lines that have been generated.
The original prompt does not appear in the final lyrics output.

\begin{wrapfigure}[23]{r}{2.25in}
    \scriptsize
    \vspace{-1.5\baselineskip}
    \begin{tabular}{|p{2.1in}|}
    \hline
\begin{enumerate}[topsep=0pt,itemsep=-1ex,partopsep=1ex,parsep=1ex,wide]
    \item Let $s_1$ and   $s_2$ be two   potentially-rhyming phoneme sequences.
    \item Replace ER with UH R in both sequences.
    \item Let $v_1$ and $v_2$ be the last stressed vowels in $s_1$ and $s_2$.
    \item Let $w_1$ and $w_2$ be last vowels in $s_1$ and $s_2$.
    \item Let $s_1 = (a_1 v_1 x_1 w_1 c_1)$. Likewise, let $s_2 = (a_2 v_2 x_2 w_2 c_2)$.
    \item Output NO under any of these circumstances:
    \begin{enumerate}[topsep=0pt,itemsep=-1ex,partopsep=1ex,parsep=1ex]
         \item $v_1 \neq v_2$
         \item $w_1 \neq w_2$
         \item \bm{$c_1 \neq c_2$}
         \item $a_1 \neq \emptyset$ and $a_2\neq \emptyset$ and $a_1 = a_2$
    \end{enumerate}
    \item If $x_1$ and $x_2$ are single phonemes:
    \begin{enumerate}[topsep=0pt,itemsep=-1ex,partopsep=1ex,parsep=1ex]
         \item {\bf If \bm{$x1\sim x_2$}, then output YES.}
         \item Otherwise, output NO.
     \end{enumerate}
    \item If $x_1$ and $x_2$ contain different numbers of vowels, output NO.
    \item Let $p_1$ and $q_1$ be the first and last phonemes of $x_1$. Let $p_2$ and $q_2$ be the same for $x_2$.
    \item {\bf If \bm{$(p_1 = p_2)$} and \bm{$(q_1 \sim q_2)$}, output YES.}
    \item {\bf If \bm{$(p_1\sim p_2)$} and \bm{$(q_1 = q_2)$}, output YES.}
    \item Otherwise, output NO.
\end{enumerate}\\
\hline
    \end{tabular}
    \caption{Near-rhyme detection algorithm~\cite{Ghazvininejad2016GeneratingTP} with modified lines highlighted.}
    \label{tab:near-rhymes}
\end{wrapfigure}
\textbf{Near-Rhyme Dictionary.}
There are plenty of perfect-rhyme dictionaries. 
A lot of music uses {\em near-rhymes}, which violate the rules of rhymes in subtle ways. 
Whatever one thinks of the artist, Eminem, he once rhymed ``discuss me'', ``disgusting'', and ``just obscene''. 
It is used quite frequently to amazing effect in rap and hip-hop.
But also Imagine Dragons, so take nothing for granted.
Anyway, figuring out whether two words are near-rhymes isn't as straight-forward as determining whether two words are perfect rhymes.
Ghazvininejad et al.~\cite{Ghazvininejad2016GeneratingTP} identified an algorithm for detecting near-rhymes, shown in
Figure~\ref{tab:near-rhymes}.
I made two changes.
The first was to delete line 6(c); I allow end consonants to be different because when lyrics are sung, the fina consonan of words are ofte softene, de-emphasize, or blende with the nex word-sound.
I also changed line 7(a), 10, and 11 where I should have used the sound of phonemes to determine similarity.
Instead, I created a set of rules to determine if two phonemes were similar. 
I found this to work better for lyrics because I had greater control of what sounded good when sung out loud.
For example, I specify a rule that phonemes with `r' components should be marked as similar to a lot of vowels. 
These design decisions gave my system a bit of a British sound, which is fine by me because I listen to a lot of pop music from the United Kingdom.\footnote{British people would probably disagree.}
I created a near-rhyme dictionary from the 20,000 most frequently used in the English language.

\textbf{Rhyme Selection.}
When the input scheme specifies, the system needs to pick a word that rhymes with one of the words in the rhyme map. 
We can pick any word from the rhyme dictionary, but how do we know a rhyming word is going to be contextually relevant?
We need a way to rank these words on relevance. 
I tried using the cosine similarity of language model contextual embeddings.
It turns out that there are some words that are just ``close'' to a lot of words and the system producing a lot of boring rhyme words.
Because the system would be using the rhyme word after some number of interstitial words that have not yet been generated,
I  needed a way of guessing if any of the candidate words would be probable in future.
The system uses GPT-2 to generate $n$ sequences of length $m$. 
As the system sample words to generate the sequences, it also measures each the rank of each rhyme candidate from the rhyme dictionary in the token distribution at each of the $m$ token positions in each of the $n$ sequences.
We finally pick the word from the rhyme dictionary with the highest average logit, which indicates that it is most likely to naturally occur during generation anyway.
I have not formally evaluated if this is significantly different from cosine similarity between candidate and a context prompt.
Anecdotally, I found that cosine similarity was resulting in more boring candidate choices and thus resulting in more boring lyrics;
the above look-ahead similarity technique seems to rank more interesting (more rare) words higher while not being too random.

\textbf{Backward Generation.}
Now that we have chosen a rhyme word to end a line,
how do we ensure that a line ends with a chosen rhyme word?
If we generate backward from that word, it will be guaranteed.
Fortunately, XLNet~\cite{Yang2019XLNetGA} can be induced to do just that.
Given the input ``\texttt{context\_1 ... context\_n MASK MASK ... MASK rhyme\_word}'', XLNet can fill each mask position one at a time.
XLNet attends to the context words at the beginning and to the rhyme word at the end when making it's decision about how to fill each mask position.
The system fills the masks starting from last and moving backward to the first, which seems to help with attending to the rhyme word.
But wait, how many mask positions should we have? 
Because tokens don't correspond to syllables, one cannot say for sure.
The system constructs a prompt with one mask and incrementally produces prompts with more masks until it has tried $\lceil s \times 1.5\rceil$ masks where $s$ is the target number of syllables for the output line. 
We have to go higher than $s$ because the language model can generate tokens corresponding to white space, punctuation, etc.
While we are generating, we disallow repeat tokens, punctuation, numbers, line breaks, or token corresponding to non-alphanumeric characters.
The system samples more tokens than necessary for each mask and iterates from the most likely to least likely, taking the first one that does not violate any of the restrictions listed above.
We repeat this entire iterative generation process $n \times 2$ times, ending the prompt with the rhyme word half the time and ending the prompt with the rhyme word followed by a period the other half.
This gives the system the option of a line that continues onto the next line or ending the line in a ``complete thought.''
Any lines that don't have the exact number of syllables are discarded, unless there are no lines with the requisite number of syllables, in which case, the system will allow lines with fewer syllables.

{\bf Forward Generation.}
When the rhyme scheme indicates that a line doesn't need to end in a rhyme, or when the line must end in a rhyme but there are no words that are yet associated with the line's rhyme index in the rhyme map (i.e., this is the first time a rhyme index is encountered) then the system uses GPT-2 to generate one token at a time.
Empirically we find that GPT-2 produces more fluent results than XLNet and is thus a better option when generating forward.
The system counts the number of syllables after each token is sampled and generation process stops after the specified number of syllables is generated.
If too many syllables are generated, the line is discarded.
This process is replicated $n$ times per line to give the system a number of options to pick from.

\textbf{Picking a Line.}
For each line in the song, either backward generation or forward generation is used and produces a number of candidate lines.
The system chooses randomly, proportional to the line's posterior probability when run through GPT-2.
One would think this would choose the most ``boring'' line, but believe me, the lines are already kind of weird. 
Mathematically it is unclear how to distinguish between a weird but fluent sequence and non-fluent garbage, so my design decision was to go with more probable lines. 
The rhyme selection process drives the interestingness by coming up with unusual words that have to be fit into the existing context of the song.
The system also has an interactive mode where the user can pick from the candidates.

\begin{wrapfigure}[20]{r}{3.0in}
\vspace{-1.0\baselineskip}
\includegraphics[width=3in]{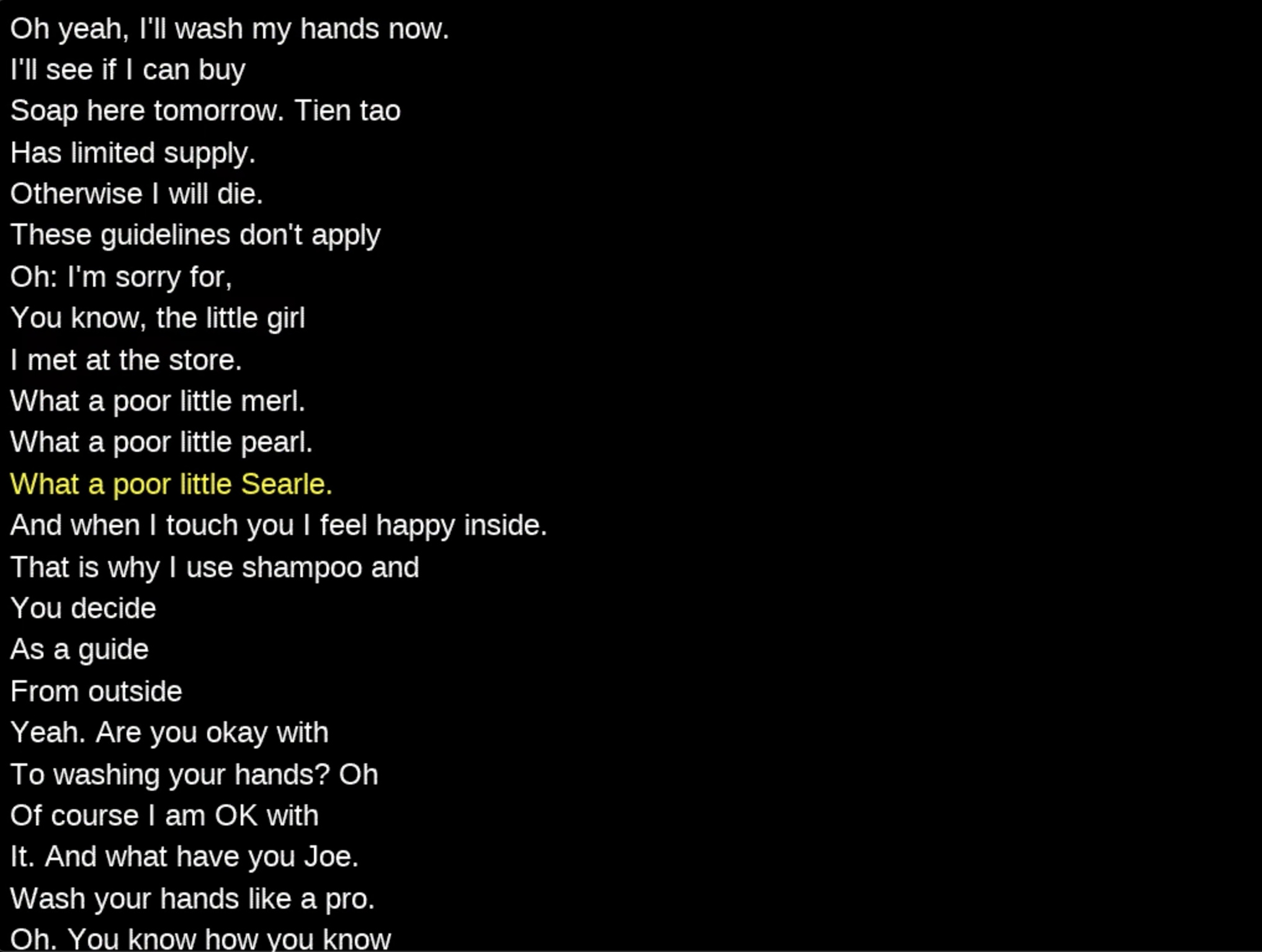}
\caption{Screenshot of the karaoke video generation, corresponding to Table~\ref{tab:weird}.
Lines are highlighted in sync with when the original lyrics would have been sung ({\em I Want to Hold Your Hand} by The Beatles).
}
\label{fig:karaoke}
\end{wrapfigure}
\textbf{Re-contextualization.}
As the number of lines grows, GPT-2 and XLNET are less likely to attend to the original prompt words and the topic can drift. 
Re-contextualization is a process whereby the the system splices the prompt text into the song lyrics after a period so that the language model generators are more likely to attend to the prompt words and generation stays on topic.
The splicing of prompt words throughout the lyrics is just for the generators and do not appear in the final output.

\textbf{Post-processing.}
The user can pre-specify a number of post-processing macros.
One command is to repeat words in a line and append it to the end of a line. 
You can see this in Table~\ref{tab:examples} in the final line of the generated output.
One can also repeat a line as a separate line. 
One can also append or prepend ``oooh'', ``aaah'', ``yeah'', ``unh'', ``whoo'', ``shamon'', or whatever exclamation is necessary.

\textbf{Karaoke Video Generation.}
What good is this if you can't sing along with instrumental music in the background?
The system is also capable of generating a karaoke video. 
Given a sound file of the melody and timing information about the original lyrics, the system creates a video that plays the music and shows each line of the newly generated lyrics at the appropriate time.
See Figure~\ref{fig:karaoke}.

\textbf{Fine-Tuning.}
One can fine-tune GPT-2 and XLNet if one wants to adopt the vocabular of a particular corpus. 
The rhyme and syllable constraints will mostly dominate the form and style of the output, but the system will prefer different words.
I generally find that careful choice of a prompt is enough to bias the vocabulary choices.

Algorithm~\ref{algo:main} shows the Weird AI Yankovic main generation loop.

\begin{algorithm}[t]
\footnotesize
\SetKw{KwIn}{in}
\SetKw{KwOr}{or}
\SetKw{KwFalse}{false}
\SetKwFunction{genRhymeLines}{generate\_rhyme\_lines}
\SetKwFunction{genTerminalNonRhymeLines}{generate\_terminal\_non\_rhyme\_lines}
\SetKwFunction{genNonRhymeLines}{generate\_non\_rhyme\_lines}
\SetKwFunction{pickBestCandidate}{pick\_best\_candidate}
\KwData{$prompt$ is a string; $scheme$ is a list of lines where each line is a list of segments where  $segment_i=\langle s_j, r_j, e_j\rangle$ for $j=1...n$ such that $s_j$ is the number of syllables in the segment, $r_j$ is rhyme index or string, and $e_j$ is an optional signifier that this segment should end a sentence; and $recontextualize?$ is a boolean.}
\KwResult{A list of strings constituting the lines of the lyrics}
 $context\gets prompt$\\
 \For{$line$ \KwIn $lines$}{
    \If{$recontextualize?={\rm true}$}{
        Insert $prompt$ in $context$ after last occurring period\\
    }
    \For{$segment$ \KwIn $line$}{
        $target\_syllables\gets$ number of syllables specified in $segment$\\
        $rhyme\_index\gets$ rhyme index specified in $segment$\\
        $end?\gets$ true if $segment$ specifies the segment ends in a period\\
        \If{$rhyme\_index$ is a string \KwOr $rhyme\_index$ \KwIn $rhyme\_map$}{
            $end\_targets\gets$ pick rhyme words or use $rhyme\_index$\\
            \For{$target$ \KwIn $end\_targets$}{
                $candidates\gets candidates +$\genRhymeLines($target$, $context$, $target\_syllables$, $end?$)\\
            }
        }
        \Else{
            \If{$end?={\rm true}$}{
                $candidates\gets candidates +$\genTerminalNonRhymeLines($context$, $target\_syllables$)\\
            }
            \Else{
                $candidates\gets candidates +$\genNonRhymeLines($context$, $target\_syllables$)\\
            }

        }
        $best\gets$\pickBestCandidate($candidates$, $context$)\\
        $context=context+best$\\
        $final\_segments\gets final\_segments + best$\\
    }
    $final\_lines = final\_lines + final\_segments$\\
  }
 \caption{The lyric generation loop.}
 \label{algo:main}
\end{algorithm}


\section{Examples}

Table~\ref{tab:hamilton} shows a number of additional examples of system output alongside the original lyrics.
If there are exact word matches between the new lyrics and the original lyrics, as in end of {\em Mad World} (third example), it is because I used the input constraints to force word choices.
All examples were first runs with a given set of inputs.

One way to measure the success of a creative AI system is the {\em curation coefficient}, the number of runs necessary before a human feels comfortable sharing a generated output~\cite{Colton2012ComputationalCT}.
That is, how many runs produce content that are not worth sharing with a public audience?
In general I find that I will get a set of lyrics that is amusing and coherent enough to share on Twitter in the first 3-5 runs. 

To get a more accurate estimate of the curation coefficient, I ran the system on the same set of inputs as many times as necessary to produce 15 sets of lyrics I would be willing to share on Twitter.com.\footnote{Lyrics were shared publicly at on my Twitter feed: \url{https://twitter.com/mark_riedl/status/1304242039337504768}} \footnote{If you are a researcher from a distant future where Twitter.com doesn't exist, please accept my apologies.}
I configured the system to rewrite the last lines of the song ``Weird Science'' by Oingo Boingo using the prompt
``I've created a monster.''
I required 31 runs, resulting a curation coefficient of $\sim$2.06.
Table~\ref{tab:weird} shows the complete set of runs, with the 15 shared runs in black and those not shared in red.
The input scheme is also given in the bottom right corner.
Since these are three-line runs, I would expect to see a higher curation coefficient for longer lyrics.
The most common reasons for rejection are (a)~linguistic disfluencies (e.g., ``An order to help you out in''), (b)~sheer jibberish (e.g., ``And my character is not
particularly throughout
of your game world''), (c)~nonsense words (e.g., ``and find out fout''), and (d)~anything even remotely racist or sexist.
I acknowledge that this experiment is not very scientific and experimenter bias may be present, which is why I give the full set of runs so one can judge for oneself whether the coefficient should be higher or lower.

We do not provide a more formal evaluation of our system.
%
However, the system was released publicly and within a day I received an email requesting a ``fart'' mode. 
Further research is required to determine if this is an effective metric for evaluating creative language generation systems.


\section{Conclusions}

Weird AI Yankovic is a demonstration that stylistic control in neural text generation can be achieved through sampling and a combination of forward and backward generation.
Whereas a lot of AI research focuses on data and novel neural model encoder and decoder architectures, we show that careful design of the sampling algorithm is an equally important part of practical and effective neural text generation. 
Being able to generate while attending to a historical context and also a word at the end of the sequence is useful for controllable text generation because it allows a system to make token-by-token decisions based on constraints regarding the end of the sequence while attending to text that comes before the masks.

One of the reasons why the Weird AI Yankovic system might be considered successful is because creative systems are designed to embrace failure.
Indeed, parody is about expressing an idea with constraint on the style and shape of the language that would not be present in ordinary everyday communication. 
The result is often awkward language usage when human do it.
Seeing the awkward solution that succeeds in meeting all the constraints provides a cathartic pleasure release in the audience.
This is accentuated in generative neural language model, which can fail in unexpected ways and non-human ways.
Indeed, one of the appeals of the Weird AI Yankovic, if there are in fact any, is seeing the train wreck of text that is output and deriving delight in the close-but-not-quite-sensical results.
Weird AI Yankovic is a case study in design trade-offs---when to handle failure modes and when to permit failures---where improvements that reduce the likelihood of failed outputs also make the outputs ``safer'' but also less interesting.


\begin{table}[t]
    \centering
    \scriptsize
    \begin{tabular}{p{1.75in}|p{1.75in}|l}
    {\bf Original Lyrics} & {\bf Generated Lyrics} & {\bf Scheme}\\
    \hline
    {\bf {\em My Shot} by Lin Manuel Miranda} & \multicolumn{2}{l}{\bf Prompt: ``I want to start a revolution''}  \\
    \hline
    I am not throwing away my shot &
    But no mean I want to start from taught & $(9, 1)$\\
    I am not throwing away my shot &
    Religious belief espoused throughout & $(9, 1)$\\
    Hey yo, I'm just like my country &
    Hey yo a society to & $(6, 2)$\\
    I'm young, scrappy and hungry &
    Blindly pursue. But no sir & $[(4, 2), (3, 2)]$\\
    And I'm not throwing away my shot &
    Give me a benefit of the doubt. & $(9, 1, {\rm :end})$ \\
    I'ma get a scholarship to King's College &
    On February & $(11, 3)$\\
    I prob'ly shouldn't brag, but dag, I amaze and astonish &
    And March we will have to Maria to our honorary & $[(6, 4), (2, 4), (7, 3)]$\\
    The problem is I got a lot of brains but no polish &
    Canons in every reverie and subsidiary & $(14, 3)$\\
    I gotta holler just to be heard &
    It is very difficult indeed & $(9, 5)$\\
    With every word, I drop knowledge &
    Not to concede. Arbitrary & $[(4, 5), (4, 3)]$\\
    I'm a diamond in the rough, a shiny piece of coal &
    And unreasonable of the just honorable & $[(7, 21), (6, 6)]$\\
    Tryin' to reach my goal my power of speech, unimpeachable &
    Dishonorable imposition on the honourable. & $[(5, 6), (10, 6, {\rm :end})]$\\
    Only nineteen but my mind is older &
    It is difficult indeed to do this. & $(10, 7)$\\
    These New York City streets get colder, I shoulder &
    But yet we have all agreed to chris' will and chris' & $[(9, 7), (3, 7)]$\\
    Every burden, every disadvantage &
    In our hearts it is only a little more & $(12, 8)$\\
    I have learned to manage, I don't have a gun to brandish &
    Certain that the coup d'or will not come or that the coup d'or & $[(6, 8), (8, 8)]$\\
    I walk these streets famished &
    Will come with a guitar & $(6, 8)$\\
    The plan is to fan this spark into a flame &
    In hand and maintained that this is more certain. & $[(2, 9), (3, 9), (2, 12), (4, 10)]$\\
    But damn, it's getting dark, so let me spell out the name &
    Sustained that this is chris' way of eliminating & $[(2, 9), (4, 12), (7, 10)]$\\
    I am the A-L-E-X-A-N-D-E-R we are meant to be. &
    Any chance of us all coming to terms with this. & $(12, 11, {\rm :end})$\\
    & & \texttt{rhyme: index $1\rightarrow {\rm shot}$}\\
    & & \texttt{post: prepend ``hey yo''}\\
    & & \texttt{~~~~~~~on line 3}\\
    \hline
    {\bf {\em Sound of Silence} by Simon \& Garfunkel} & \multicolumn{2}{l}{\bf Prompt: ``Hello darkness, my old friend.''} \\
    \hline
    Hello darkness, my old friend &
    But Darkness never explained. & $(7, 1)$ \\
    I've come to talk with you again &
    But once gained, Darkness never gained. & $(8, 1, {\rm :end})$\\
    Because a vision softly creeping &
    My shadow has vanished from & $(8, 2)$\\
    Left its seeds while I was sleeping &
    This plane of existence. See em. & $(8, 2)$\\
    And the vision that was planted in my brain &
    On the horizon and look at Darkness in & $(11, 3)$\\
    Still remains &
    The between. & $(3, 3, {\rm :end})$\\
    Within the sound of silence &
    See em. In the silence. & $(6, {\rm silence}, {\rm :end})$\\
    \hline
    {\bf {\em Mad World} by Tears for Fears} & \multicolumn{2}{l}{\bf Prompt: ``All around me are familiar faces. Mad world.''} \\
    \hline
    All around me are familiar faces & 
    Of course I do remember them very & $(10, 1)$\\
    Worn out places, worn out faces & 
    The contrary. Old Jack Kerry. & $[(4, 1), (4, 1, {\rm :end})]$\\
    Bright and early for their daily races & 
    The Henry factory. Wilmer's diary. & $(10, 1)$\\
    Going nowhere, going nowhere & 
    In my mind the image of the. & $[(4, 2), (4, 2, {\rm :end})]$\\
    Their tears are filling up their glasses & 
    Is one of them which my mother & $(9, 3)$\\ 
    No expression, no expression & has
of me and my father will die. & $[(4, 4), (4, 4, {\rm :end})]$\\
    Hide my head I want to drown my sorrow & 
    Are not so different from them but one & $(10, 5)$\\
     No tomorrow, no tomorrow & 
     Hamburger bun. They are the un. & $[(4, 5), (4, 5, {\rm :end})]$\\
    & \\
    And I find it kind of funny & 
    Or something like that. Its funny. & $(8, {\rm funny})$ \\
    I find it kind of sad &
    But very strange. Its sad. & $(6, {\rm sad}, {\rm :end})$\\
    The dreams in which I'm dying are the best I've ever had & 
    Just never know what other people think of me. And Brad. & $(14, 6)$\\
    I find it hard to tell you 'cause I find it hard to take & 
    On another hand its not just what they say but how & $(13, 7)$\\
    When people run in circles it's a very, very & 
    With other people their opinions are those things & $(13, 8)$\\
    Mad world, mad world & 
    Mad world, mad world. & $(4, {\rm Mad~world, mad~world})$\\
    & & \texttt{rhyme: index $6\rightarrow {\rm sad}$}\\
    \hline
    {\bf {\em Can't Touch This} by M.C. Hammer} & \multicolumn{2}{l}{\bf Prompt: ``You cannot touch this.''} \\
    \hline
    My, my, my my music hits me so hard &  And and and and should not touch this again & $[(1, {\rm None}), (6, 1)]$\\
    Makes me say, ``Oh my Lord'' & Do not touch to the tongue. & $(3, 1, {\rm :end})$\\
    Thank you for blessin' me & The person in question & $(6, 2)$\\
    With a mind to rhyme and two hype feet & And no one trying indigestion & $[(3, 3), (2, 3), (4, 2, {\rm :end})]$\\
    It feels good, when you know you're down & Do it in anger because they & $(8, 4)$\\
    A super dope homeboy from the Oaktown & Should not touch the tongue say it ye and yea. & $(10, 4, {\rm :end})$\\
    And I'm known as such & I said unto you & $(5, 5)$\\
    And this is a beat, uh, you can't touch & And if anything, uh, you pursue. & $[(5, {\rm None}), (3, 5, {\rm :end})]$\\
    & & \texttt{post: repeat 1st word $3\times$},\\
    & & \texttt{~~~~~~~insert ``uh'' line 8}\\
    \hline
    \end{tabular}
    \caption{Example first runs with different input constraints and prompts.}
    \label{tab:hamilton}
\end{table}


\section{Societal Implications}

As a neural language generation system, our system faces the same potential pitfalls as other contemporary language learning systems. 
Namely, it is prone to echoing the biases present in the dataset, including
{\em prejudicial} biases or descriptions of non-normative behavior.
Prejudicial biases are biases that are unwanted because they result in language that demeans or infers hatred or violence toward certain people.
Non-normative behavior~\cite{peng} is that which is not considered outside the norms of a particular group of people. 
In the United States this might include descriptions of public sex or murder.
The system as described in it's current manifestation has little practical value and thus presents little chance of harm at scale.
As with any broad capability technology, it can be put to purposes that are benign, malicious, or negligent that have not been envisioned by the author.

One general concern in artificial intelligence is the prospect of automating jobs. 
Unlike mundane tasks where there is little opportunity for human improvisational problem solving, ``creative'' AI systems introduce the prospect of more specialized forms of work currently believed to be uniquely considered only possible by humans.
Weird AI Yankovic is not good enough to completely replace human musical artists.
Part of the reason is that GPT-2 and XLNet, while reasonably good at producing fluent language, do not have any particular understanding of what they are generating or how the resultant language will impact a human audience.
Systems such as GPT-3 may make significant gains in fluency but are theoretically limited with respect to its ability to generate text with an intended impact on the reader/audience.
However, the above discussion assumes that human consumers of music---or any type of creative expression---will accept AI-generated creative content, even if it of equal objective quality to human-generated content.
One of the reasons we value creative expression is because of the tacit acknowledgement of the effort that went into the creative expression. 
I hypothesize that we will not value computational effort as equivalent to human effort---humans must make tradeoffs on how they spend their finite time and resources in a way whereas computational effort is cheap.

One use that this work may be put to is human-AI co-creation tool, wherein a human works with a computational system in a mixed-initiative interaction.
Although Weird AI Yankovic provides an interactive mode where a user can manually select candidate lines.
However, this does not align with actual needs or creative processes of human artists.
There is a common misconception that any fully autonomous creative system can be simply folded into a mixed-initiative framework to make an AI creativity support.
Kristen Stewart, of the Twighlight movie fame, wrote a paper documenting how difficult it is to use AI creative systems that were not designed for co-creation when trying to achieve a desired aesthetic solution~\cite{Joshi2017BringingIT}.
Weird AI Yankovic in interactive mode has more in common with a party game than a creativity support tool.

Finally, one societal impact of Weird AI Yankovic is that one may accused of copyright infringement. 
Despite that fact that parody is protected in the United State of America as fair use, this does not stop those with vested interested in the copyright of an original work of music using tactics to limit parody. 
The  International Federation of the Phonographic Industry (IFPI), which manages copyright protection for a number of music artists, including Michael Jackon's estate, sent Twitter.com a take-down request targeting one om tweets because it contained a link to a karaoke video generated by the Weird AI Yankovic system.
There is a legal gray zone wherein the lyrics, which are textually overlaid on a black screen (see Figure~\ref{fig:karaoke} for an example) while an original instrumental track plays.
Can one derive value from listening to the instrumental track without reading the parody lyrics and thus deprive a copyright holder from monetary value?
Further complicating the issue, fair use laws are not uniform internationally, meaning that generated lyrics may be protected in some countries but not others.

\begin{table}[t]
    \centering
    \scriptsize
    \begin{tabular}{p{1.6in}|p{1.6in}|p{1.6in}}
    \multicolumn{3}{l}{\bf Curated runs}\\
    \hline
    For my own personal use & And don't forget that i'm just & I'll have it in the hands of\\
    Only to reintroduce & Constantly chasing the guste. & My eternally belove\\
    That beast to the ooh, weird science & To prove my point ooh, weird science & Waswolf boyfriend ooh, weird science\\
    \hline
    Have to make him do something & For those of you who don't know & When he went missing last week\\
    Bad like this twentysomething & Exactly what molding dough & He was surrounded by chic\\
    A couple of ooh, weird science & Of that molder ooh, weird science & Was going to ooh, weird science\\
    \hline
    Are you really that big with & I will kill it and live on & And i know how to control.\\
    Other people messing with & A hundred million yuan. & And he started with rock'n'roll.\\
    Just a little ooh, weird science & In the meantime ooh, weird science & I want you to ooh, weird science\\
    \hline
    And this monster will only & The first one is called the black & And now he'll come after us\\
    Ever be sad and lonely. & Cat but he is a chirac. & All and then we never guess\\
    You must defeat ooh, weird science & It is called the ooh, weird science & And the next time ooh, weird science\\
    \hline
    To make her feel good about & And now we are ready to & Think about all those guys who\\
    Having the thing she had sought. & Step onboard the ship that threw & Never get seasonal flu\\
    Would love to see ooh, weird science & Canisters out ooh, weird science & Had so many ooh, weird science\\
    \hline
    \multicolumn{3}{l}{\bf Rejected runs}\\
    \hline
    \textcolor{red}{And we are all living in} & \textcolor{red}{And i have created two.} & \textcolor{red}{And here we go again with}\\
    \textcolor{red}{The world we are living in.} & \textcolor{red}{And those two have not been threw} & \textcolor{red}{Without the monster and with}\\
    \textcolor{red}{So if you're like ooh, weird science} & \textcolor{red}{And will not throw ooh, weird science} & \textcolor{red}{Would be nothing ooh, weird science}\\
    \hline
    \textcolor{red}{You i will be a giant.} & \textcolor{red}{Of my own creation by} & \textcolor{red}{And i don't really care if}\\
    \textcolor{red}{You that is what i wiant.} & \textcolor{red}{Simply using the cute shy} & \textcolor{red}{This might be written by ziff.}\\
    \textcolor{red}{And it will be ooh, weird science} & \textcolor{red}{And the little ooh, weird science} & \textcolor{red}{Has also done ooh, weird science}\\
    \hline
    \textcolor{red}{Is a simple program which} & \textcolor{red}{Have created some monster} & \textcolor{red}{That you can use on your own}\\
    \textcolor{red}{Can helps you create a fritch} & \textcolor{red}{More like thesanta santa} & \textcolor{red}{Side and also in a sloan.}\\
    \textcolor{red}{Of different ooh, weird science} & \textcolor{red}{Are more like that ooh, weird science} & \textcolor{red}{Would also have ooh, weird science}\\
    \hline
    \textcolor{red}{And my character is not} & \textcolor{red}{And we are going to do} & \textcolor{red}{Have you ever seen that one}\\
    \textcolor{red}{Particularly throughout} & \textcolor{red}{About with what they pursue.} & \textcolor{red}{Little black terrebonne}\\
    \textcolor{red}{Of your game world ooh, weird science} & \textcolor{red}{Also it will ooh, weird science} & \textcolor{red}{Would grow into ooh, weird science}\\
    \hline
    \textcolor{red}{I'm here to help you out in} & \textcolor{red}{And i'm gonna try it out} & \textcolor{red}{If people are afraid that}\\
    \textcolor{red}{An order to help out in} & \textcolor{red}{There to see and find out fout} & \textcolor{red}{The inmarsat inmarsat}\\
    \textcolor{red}{And out of this ooh, weird science} & \textcolor{red}{For myself if ooh, weird science} & \textcolor{red}{Have something to ooh, weird science}\\
    \hline
    \textcolor{red}{But let me tell you something} & \multicolumn{2}{c}{$(7, 1)$}\\
    \textcolor{red}{About this twentysomething} & \multicolumn{2}{c}{$(7, 1)$}\\
    \textcolor{red}{That came out of ooh, weird science} & \multicolumn{2}{c}{$[(4, {\rm None}), (4,$ ``ooh, weird~science''$)]$}\\
    \hline
    \end{tabular}
    \caption{Fifteen curated runs of Weird AI Yankovic generating the final lines of ``Weird Science'' by Oingo Boingo using the scheme in the lower right corner.
    }
    \label{tab:weird}
\end{table}

\bibliography{neurips_2020.bib}
\bibliographystyle{abbrvnat}

\end{document}